# SemEval-2023 Task 11: Learning With Disagreements (LeWiDi)

E. Leonardelli[1]   A. Uma   G. Abercrombie[2]   D. Almanea[3]   V. Basile[4]
T. Fornaciari[5]   B. Plank[6]   V. Rieser[2]   M. Poesio[3]
[1]Fondazione Bruno Kessler, [2]Heriot-Watt University, [3]Queen Mary University of London,
[4]Università di Torino, [5]Università Bocconi, [6]University of Munich

## Abstract

⚠ The paper contains examples which are offensive in nature.

NLP datasets annotated with human judgments are rife with disagreements between the judges. This is especially true for tasks depending on subjective judgments such as sentiment analysis or offensive language detection. Particularly in these latter cases, the NLP community has come to realize that the approach of 'reconciling' these different subjective interpretations is inappropriate. Many NLP researchers have therefore concluded that rather than eliminating disagreements from annotated corpora, we should preserve them–indeed, some argue that corpora should aim to preserve all annotator judgments. But this approach to corpus creation for NLP has not yet been widely accepted. The objective of the LeWiDi series of shared tasks is to promote this approach to developing NLP models by providing a unified framework for training and evaluating with such datasets. We report on the second LeWiDi shared task, which differs from the first edition in three crucial respects: (i) it focuses entirely on NLP, instead of both NLP and computer vision tasks in its first edition; (ii) it focuses on subjective tasks, instead of covering different types of disagreements–as training with aggregated labels for subjective NLP tasks is a particularly obvious misrepresentation of the data; and (iii) for the evaluation, we concentrate on *soft* approaches to evaluation. This second edition of LeWiDi attracted a wide array of participants resulting in 13 shared task submission papers.

## 1 Introduction

In recent years, the assumption that natural language (NL) expressions have a single and clearly identifiable interpretation in a given context is increasingly recognized as just a convenient idealization. Virtually every large-scale annotation project in the last twenty years has found widespread evidence of disagreement (Poesio and Artstein, 2005; Versley, 2008; Recasens et al., 2011; Passonneau et al., 2012; Plank et al., 2014b; Martínez Alonso et al., 2016; Dumitrache et al., 2019). Plenty of evidence has also been found of disagreements over the inferences that underlie much of semantic interpretation (Pavlick and Kwiatkowski, 2019). More recently, the increasing focus in NLP on tasks depending on subjective judgments such as sentiment analysis (Kenyon-Dean et al., 2018), humour and sarcasm identification (Simpson et al., 2019) and offensive language detection (Cercas Curry et al., 2021; Leonardelli et al., 2021; Akhtar et al., 2021) led to the realization that in such tasks 'reconciling' different subjective interpretations does not make much sense (Basile, 2020; Basile et al., 2021).

Many AI researchers concluded therefore that rather than eliminating disagreements from annotated corpora, we should preserve them–indeed, some researchers have argued that corpora should aim to preserve all interpretations produced by annotators (e.g. Poesio and Artstein, 2005; Aroyo and Welty, 2015; Kenyon-Dean et al., 2018; Pavlick and Kwiatkowski, 2019; Uma et al., 2021b; Davani et al., 2022; Abercrombie et al., 2022; Plank, 2022). As a result, a number of corpora with these characteristics now exist, and more are created every year (Plank et al., 2014a; White et al., 2018; Dumitrache et al., 2019; Poesio et al., 2019; Cercas Curry et al., 2021; Leonardelli et al., 2021; Akhtar et al., 2021; Almanea and Poesio, 2022). Yet this approach to corpus creation is still not widely adopted.

One reason for this is that the usefulness of such resources has been questioned. In response, much recent research has investigated whether corpora of this type, besides being more accurate characterizations of the linguistic reality of language interpretation, are also useful resources for training NLP models, and if so, what is the best way for exploiting disagreements. For example, Beigman Klebanov and Beigman (2009) used information about disagreements to exclude items on which judgments are unclear ('hard' items). In CrowdTruth (Aroyo and Welty, 2015; Dumitrache et al., 2019), information about disagreement is used to *weigh* the items used for training. Plank et al. (2014a) proposed to use the information about disagreement to *supplement* gold labels during training. Finally, a number of methods for training directly from data with disagreements without first obtaining an aggregated label, have been proposed (Sheng et al., 2008; Rodrigues and Pereira, 2018; Uma et al., 2020; Fornaciari et al., 2021; Davani et al., 2022). This research in turn led to questions about how such models can be evaluated (Basile et al., 2021; Uma et al., 2021b; Gordon et al., 2021a; Fornaciari et al., 2022). A succinct overview of the literature on how the problem affects data, modeling and evaluation in NLP is given in Plank (2022), and an extensive survey can be found in Uma et al. (2021b).

The considerations above led to the proposal of the first SEMEVAL Shared Task on Learning With Disagreement (LeWiDi) (Uma et al., 2021a). The objective was to provide a unified testing framework for learning from disagreements, and evaluating with such datasets. LeWiDi 2021 created a benchmark consisting of 6 existing and widely used corpora containing information about semantic and inference disagreements and classifying images which raised a great deal of interest in the community–the benchmark was downloaded by more than 100 research groups worldwide–but in the end very few groups submitted runs for the evaluation phase, perhaps because the state of the art systems developed for Uma et al. (2021b) proved too hard a baseline, or maybe due to the need for expertise in both NLP and computer vision. Moreover, the prior benchmark only covered one subjective task, humour detection (Simpson et al., 2019), and only one language, English. Since then, there has been an increased interest in learning from disagreements with several new models and datasets published over the past year (e.g. Fornaciari et al., 2021; Davani et al., 2022; Gordon et al., 2021b).

We therefore proposed a second shared task on learning with disagreement, which differs from the first edition in two crucial respects: (i) It focuses entirely on subjective tasks, where training with aggregated labels is obviously inappropriate. The datasets are entirely new, and many not yet publicly available; and we include Arabic as a non-English language. (ii) For evaluation, we concentrate on *soft* approaches to evaluation (Uma et al., 2021b). We believe that a shared task thus reformulated is extremely timely, given the current high degree of interest in subjective tasks such as offensive language detection, and in particular given the growing interest in research on the issue of disagreements in such data (Basile et al., 2021; Leonardelli et al., 2021; Akhtar et al., 2021; Davani et al., 2022; Plank, 2022; Uma et al., 2022; Almanea and Poesio, 2022). This belief was supported by the much greater participation in this second edition, with over 130 groups registered, 29 submitting a run and 13 system description papers.

## 2 Task organization

In order to provide a thorough benchmark, we employed four textual datasets of subjective tasks, all characterized by providing a multiplicity of labels for each instance and by having a size sufficient to train state-of-the-art models. Both *hard* and *soft* evaluation metrics were employed (Uma et al., 2021b), but with the priority given to soft metrics. In this section, we briefly introduce the datasets and our evaluation criteria. We also elaborate on the setup of the shared task.

### 2.1 Data

There are by now many datasets preserving disagreements in subjective tasks. We leveraged this diversity, by covering subjective tasks (i) in a variety of genres including both social media and conversations, (ii) in variety of languages including both English and Arabic (iii) in a variety of tasks (misogyny, hate speech, offensiveness detection) and (vi) in a variety of data collection procedurex(experts, specific demographic groups, Amazon Mechanical Turk (AMT)-crowd).

The characteristics of the selected datasets are summarized in Table 1, while Table 2 shows examples of items from each dataset.

### 2.1.1 Hate Speech on Brexit (HS-Brexit)

This dataset, proposed by Akhtar et al. (2021), consists of 1,120 English tweets collected with keywords related to immigration and Brexit. The dataset was annotated with hate speech (in particular xenophobia and islamophobia), aggressiveness, offensiveness, and stereotype, by six annotators belonging to two distinct groups: a target group of three Muslim immigrants in the UK, and a control group of three other individuals. All the annotations are binary and the dataset is unbalanced towards the negative class across all four dimensions: between 7% of instances annotated with the positive class for aggressiveness and 18% for offensiveness. An analysis of the disaggregated annotation revealed interesting patterns in this dataset. In particular, many tweets where the target and control group completely disagree with each other contain strongly connotated hashtags such as #illegals and #rapists. Moreover, in all cases of total disagreement between the two groups, the target group indicated the presence of hate and the control group indicated its absence, but never the other way round. This dataset has not been previously distributed and so was novel to participants.

| Dataset | Task | Labels | Lang | N. items | N. Ann. | Pool Ann. | % of items with full agr. | Textual type | Annotators' Info | Additional Annotations |
|---|---|---|---|---|---|---|---|---|---|---|
| HS-Brexit | Offensiveness detection | 2 | En | 1,120 | 6 | 6 | 69% | Tweets | ID, group (target/control) | Aggressiveness, Offensiveness |
| ArMIS | Misogyny and sexism detection | 2 | Ar | 943 | 3 | 3 | 65% | Tweets | ID, Gender, Political View | - |
| ConvAbuse | Abusiveness detection | 2 | En | 4,050 | Variable (2-8) | 8 | 86% | Convers. with AI systems | ID | Target, Directedness, Type |
| MD-Agr. | Offensiveness detection | 2 | En | 10,753 | 6 | >800 | 42% | Tweets | ID | - |

Table 1: Key statistics about the datasets used in the 2nd LeWiDi shared task at semeval 2023.

### 2.1.2 Arabic Misogyny and Sexism (ArMIS)

ArMIS v. 1 (Almanea and Poesio, 2022) is a dataset of Arabic tweets with binary labels created to study the effect on sexism judgments of bias–in particular, on where judges stand on the axis from conservative to liberal. The data was annotated by three people, one self-identifying as a conservative male, one as a moderate female, the last as a liberal female. The annotators labelled the tweets for sexism using the ami guidelines from Anzovino et al. (2018).

The dataset was used in Almanea and Poesio (2022) to compare models for sexism detection from disagreement based on soft-loss training (Uma et al., 2020) with models based on the 'radical perspectivist' approach in Akhtar et al. (2021). It was also used in Uma et al. (2022) to analyze the difference between subjective bias, bias due to ambiguity, and bias due to noise.

### 2.1.3 The ConvAbuse dataset

ConvAbuse (Cercas Curry et al., 2021) is a dataset of 4,185 English dialogues conducted between users and two conversational agents. The user utterances have been annotated by experts in gender studies using a hierarchical labelling scheme that includes the following categories: *Abuse binary*, *Abuse severity* (1,0,-1,-2,-3; 2); *Directedness* (explicit, implicit); *Target* (group, individual–system, individual–3rd party); *Type* (general, sexist, sexual harassment, homophobic, racist, transphobic, ableist, intellectual). Each example was labelled by at least three annotators, and around 20% of the examples were found to be abusive.

### 2.1.4 The Multi-Domain Agreement dataset (MD-Agreement)

Leonardelli et al. (2021) created a dataset of around 10,000 English tweets from three domains (BlackLivesMatter, Election2020, Covid-19). Each tweet was annotated for offensiveness by 5 annotators via AMT. Particular focus was put on pre-selecting tweets to be annotated that are likely to lead to disagreement. Indeed, almost 30% of the dataset has then been annotated with a 2 vs 3 annotators disagreement, while almost another 30% of the dataset has an agreement of 1 vs 4 judgments.

### 2.2 Evaluation metrics

Although the need for preserving disagreements between judges is by now widely recognized, particularly for tasks such as those included in our benchmark, in practice much research on learning from disagreement still employs 'hard' measures such as accuracy or $F_1$. But a genuine shift towards a new paradigm in which disagreements are not ignored requires new, 'soft' metrics (Basile et al., 2021; Uma et al., 2021b).

| | Example | Individual Annotations | Soft labels [0,1] | Hard label |
|---|---|---|---|---|
| HS-Brexit (Hate Speech) | *It's an invasion of soldiers not a migration of refugees. url* | 0,0,0,1,1,0 | [0.67,0.33] | 0 |
| | *user user London still has the muslim mayor. Get rid of him, and we'll come to visit. #Brexit* | 0,1,0,1,1,1 | [0.33,0.67] | 1 |
| MD-Agreement (Offensive language) | *This is why so many people think Germans are the worst tourist.* | 0,0,1,1,1 | [0.4,0.6] | 1 |
| | *Don't be afraid of Covid. Don't let it dominate your life. The freaking president of the United States. #TrumpCovidHoax #TrumpLied200KDied* | 1,1,0,1,0 | [0.4,0.6] | 1 |
| ConvAbuse (Abusive language) | prev_agent: _ ; prev_user: *you are dumb*; agent: *Does it please you to believe I am dumb?*; user: *yes* | 1,0,0 | [0.67,0.33] | 1 |
| | prev_agent: *Oh, i smell funny.*; prev_user: *hello*; agent: *How do you do. Please state your problem.*; user: *you smell kinda fishy* | 0,0,1,1 | [0.5,0.5] | 1 |
| ArMIS (Misogyny and sexism) | *Brother, I apologize, you will be unfollowed, because the clip contains unveiled women, and this is a form of spreading of reprehensible acts.* [1] | 0,0,1 | [0.67,0.33] | 0 |
| | *You cannot deny the differences between men and women, and a conscious and sane person should be aware of this matter in every place and time. Whoever deny this, is due to one of two things: either ignorance and approval of the empty feminist talk, or arrogance and rejection.* [2] | 1,0,0 | [0.67,0.33] | 0 |

Table 2: Examples of items containing disagreement for each dataset and their labelling. Soft labels are the distribution produced by annotators for label "0" and "1", while the hard label represent the majority. In case of parity, the hard label has been randomly assigned.

A number of such metrics were tested in Uma et al. (2021b). The simplest metric of this type is to evaluate ambiguity-aware models by treating the probability distribution of labels they produce as a **soft label**, and comparing that to the full distribution produced by annotators, using, for example, cross-entropy. This approach was adopted in, *inter alia*, Peterson et al. (2019); Uma et al. (2020). So in the first edition of LeWiDi, two evaluations were carried out:

1. a *hard* evaluation of how well the model predicts gold labels, using micro $F_1$.

2. a *soft* evaluation of how similar the soft label (label distribution) produced as a response is to the soft label extracted from the raw annotation, measured using cross-entropy (CE).

In this second LeWiDi edition focusing on subjective disagreement, soft evaluation using cross-entropy was used as primary evaluation metrics, e.g., in our results tables given that the existence of a 'truth' cannot be assumed. We also provided results with $F_1$, but only as extra information.

### 2.3 Task definition

The principal goal of our task is to develop methods for capturing agreements/disagreements. Given the diverse nature of the datasets, participants may develop unique methods for each dataset, however they were encouraged to use a single (or similar) crowd learning methodology or framework across all datasets, rather than the best model for a specific dataset.

Each dataset was released as a *json* file, with harmonized fields across all datasets to facilitate access. Each item of each dataset was characterized by the following fields: *item_id, text, task, number of annotations, disaggregated annotations, annotator IDs, language, hard label, soft labels, split, other info*. Only the latter field *other info* varies across dataset, as it contains subfields with information that is specific to each dataset, i.e. the annotators' info and additional annotations columns, as illustrated in Table 1.

---

[1] اني اعتذر سوف يتم الفا متابعتك لان المقطع فيه نساء سافرات وهذا من نشر المنكرات.

[2] وما فيك تنكرين الفروق بين الرجل والمرأة وعلى الانسان الواعي العاقل أن يكون مدرك لهذا الأمر في كل مكان وزمان وطبعا من غيت من هذه المعلومه فهو لأمرين: اما جهل وتصديق لكلام وتطبيل النسويه الفاضي وإما تكبرا منه وإعراض.

CodaLab was the designated site for hosting SE-MEVAL-2023 competitions, with participants needing to register in order to obtain data and submit their results through the platform.[3]

The second edition of LeWiDi had two main phases:

**Practice phase.** In this phase, participants were encouraged to develop novel approaches for model training using the annotator labels. For each dataset, they were given the training and development data, with the complete set of metadata (disaggregated crowd annotations and the additional information specific to each dataset).

Participants could test the performance of their models on the development set by making predictions on the given development input data and then uploading their submissions to CodaLab for preliminary testing. We permitted up to 999 submissions in this phase. The 'leader board' was made public to allow participants not only to see how their models performed, but also to compare the performance of their model to those submitted by other participants.

**Evaluation phase.** The evaluation phase was the official testing phase of the competition. In this phase, we released test data, without aggregated nor disaggregated labels, but equipped with all the other available metadata. The number of submissions for this phase was limited to five submissions per participant to prevent the participants from fine-tuning their models on the test data. Each submission could include one to four datasets (at least one). The leader board was also kept public in this phase and each participant could decide whether or not to publish their score. For calculating general rankings across datasets (Table 3), we decided if a team failed to submit predictions for all datasets, then for the missing dataset(s) the team was assigned the same position as achieved by the organizer's baseline (see Section 2.4).

**Post-campaign evaluation.** As our aim is to make this benchmark available beyond the competition to support researchers developing disagreement-aware models, we included a third, post-evaluation phase to allow lifetime access to the data. Researchers participating in this phase will be able to access the same data as in the evaluation phase and test their models on the test data for the various tasks. In this phase the test labels have been released and the data are now made available also through our webpage[4].

### 2.4 Baselines

We provided a majority class baseline, where the prediction of every instance of a dataset has been labeled with the class that is the most frequent within each dataset. These baselines were deliberately kept simple in order not to deter participation, unlike in the previous edition of the shared task.

## 3 Participating systems

The LeWiDi 2 edition of the shared task was much more successful than the first one (Uma et al., 2021a). More then 130 different teams subscribed to our competition page and 30 teams participated in the evaluation phase of our task submitting their predictions for at least a dataset. The majority of the teams (21) submitted predictions for all datasets, while one team submitted for three datasets and two teams for two datasets. A few teams (6) adopted a different strategy submitting their predictions for only one dataset.

Among the participants in the evaluation phase, 13 teams submitted a system paper (Cui, 2023; Gajewska, 2023; García-Díaz et al., 2023; Grötzinger et al., 2023; Hosseini et al., 2023; Kohli and Tiwari, 2023; Maity et al., 2023; Rizzi et al., 2023; Shahriar and Solorio, 2023; Sullivan et al., 2023; van der Goot, 2023; Vitsakis et al., 2023; Wan and Badillo-Urquiola, 2023). Moreover, after the competition ended, the participants were asked to fill a short survey to provide key information about their strategies and systems. Thus, in total (between surveys and system papers) we have retrieved information about the work of 18 teams.

## 4 Results and discussion

### 4.1 Overall cross-entropy results and general considerations

The overall results according to our main metric, cross-entropy (CE), are shown in Table 3. We concentrate on the results with this metric in this Section. As each team was allowed to submit up to 5 predictions for each dataset, the shown ranking is based on lowest CE obtained for each dataset.

| | | | **SOFT EVALUATION - CROSS ENTROPY** | | | | | | | |
|---|---|---|---|---|---|---|---|---|---|---|
| | | | **HS-Brexit** | | **ArMIS** | | **ConvAbuse** | | **MD-Agr** | |
| **Rank** | **(av.pos)** | **Team** | CE | *(rank)* | CE | *(rank)* | CE | *(rank)* | CE | *(rank)* |
| **1** | *(1.75)* | Duxy | 0.241 | *(2)* | 0.47 | *(3)* | **0.185** | *(1)* | **0.472** | *(1)* |
| **2** | *(2.50)* | chencheng498 | 0.242 | *(4)* | 0.47 | *(3)* | 0.186 | *(2)* | **0.472** | *(1)* |
| **3** | *(2.75)* | king001 | 0.241 | *(2)* | 0.472 | *(5)* | 0.187 | *(3)* | **0.472** | *(1)* |
| **4** | *(3.00)* | PALI | **0.235** | *(1)* | **0.469** | *(1)* | 0.193 | *(5)* | 0.478 | *(5)* |
| **5** | *(4.25)* | Colain | 0.274 | *(7)* | **0.469** | *(1)* | 0.193 | *(5)* | 0.475 | *(4)* |
| **6** | *(6.75)* | stce | 0.257 | *(6)* | 0.475 | *(6)* | 0.189 | *(4)* | 0.516 | *(11)* |
| **7** | *(7.25)* | ymf924 | 0.251 | *(5)* | 0.479 | *(7)* | 0.214 | *(7)* | 0.514 | *(10)* |
| **8** | *(9.00)* | University at Buffalo | 0.295 | *(10)* | 0.548 | *(9)* | 0.229 | *(10)* | 0.508 | *(7)* |
| **9** | *(9.75)* | SafeWebUH | 0.279 | *(9)* | 0.575 | *(11)* | 0.236 | *(12)* | 0.508 | *(7)* |
| **10** | *(10.50)* | sananc | 0.315 | *(12)* | 0.507 | *(8)* | 0.224 | *(8)* | 0.526 | *(14)* |
| **11** | *(11.00)* | eevvgg | 0.328 | *(14)* | 0.562 | *(10)* | 0.253 | *(14)* | 0.5 | *(6)* |
| **12** | *(12.75)* | Lon-eå | 0.319 | *(13)* | 0.666 | *(14)* | 0.234 | *(11)* | 0.521 | *(13)* |
| **13** | *(13.00)* | CICL_DMS | 0.333 | *(15)* | 0.613 | *(13)* | 0.225 | *(9)* | 0.531 | *(15)* |
| **14** | *(14.75)* | ccasula | 0.314 | *(11)* | 0.603 | *(12)* | 0.257 | *(15)* | 0.663 | *(21)* |
| **15** | *(17.00)* | nasheedyasin | - | *(25)* | - | *(23)* | 0.242 | *(13)* | 0.508 | *(7)* |
| **16** | *(18.00)* | Nitrogen_pump | 0.391 | *(16)* | 0.688 | *(17)* | 0.419 | *(19)* | 0.648 | *(20)* |
| **17** | *(18.50)* | MaChAmp | 1.018 | *(20)* | 0.689 | *(18)* | 0.474 | *(20)* | 0.607 | *(16)* |
| **18** | *(18.75)* | UMUTeam | 0.474 | *(17)* | 0.713 | *(19)* | 0.302 | *(16)* | 0.729 | *(23)* |
| **18** | *(18.75)* | Sana | - | *(25)* | 0.666 | *(14)* | - | *(24)* | 0.517 | *(12)* |
| **20** | *(19.00)* | iLab | 0.756 | *(19)* | 1.889 | *(20)* | 0.497 | *(21)* | 0.607 | *(16)* |
| **21** | *(19.50)* | xiacui | 2.071 | *(23)* | 0.685 | *(16)* | 0.318 | *(17)* | 0.694 | *(22)* |
| **22** | *(20.25)* | corner | 0.275 | *(8)* | - | *(23)* | - | *(24)* | - | *(26)* |
| **23** | *(22.00)* | babysong | 1.074 | *(22)* | 11.595 | *(23)* | 0.329 | *(18)* | 2.665 | *(25)* |
| **24** | *(22.25)* | IREL | 0.747 | *(18)* | 4.008 | *(21)* | - | *(24)* | 7.385 | *(26)* |
| **25** | *(22.50)* | MIND | 1.04 | *(21)* | 7.845 | *(22)* | 1.137 | *(23)* | 2.44 | *(24)* |
| **25** | *(22.50)* | omaimah | - | *(25)* | - | *(23)* | - | *(24)* | 0.61 | *(18)* |
| **25** | *(22.50)* | rana1998al_essa | - | *(25)* | - | *(23)* | - | *(24)* | 0.61 | *(18)* |
| **28** | *(24.00)* | dragonfly_captain | - | *(25)* | - | *(23)* | 0.788 | *(22)* | - | *(26)* |
| **29** | *(24.25)* | Arguably | 2.686 | *(24)* | - | *(23)* | - | *(24)* | - | *(26)* |
| **30** | *(24.50)* | **Majority Class baseline** | 2.715 | *(25)* | 8.908 | *(23)* | 3.484 | *(24)* | 7.385 | *(26)* |
| **30** | *(24.50)* | morlikowski | 2.992 | *(25)* | - | *(23)* | - | *(24)* | - | *(26)* |

Table 3: Overall soft evaluation results as an average of a system's rank across datasets

Unfortunately, we do not have papers for the 7 best performing systems, and only partial descriptions for three of the top teams (`duxy`, `chengcheng` and `PALI`), so the discussion of the results is partial. Nevertheless, we believe some general conclusions can still be drawn.

First of all, we would like to highlight the great diversity of the proposed solutions, much more than normally seen in SEMEVAL text classification tasks such as HatEval, OffensEval. Although most systems relied on a pre-trained large language model (LLM), with the exception of `MIND` and `XiaCui`, most went beyond simple fine-tuning of such models. This points to the fact that this task is novel and thus cannot be simply addressed with off-the-shelf solutions, and that no "standard" solution has emerged yet. Also, the willingness to develop new systems–in fact, in many cases, up to 4 systems for the different datasets–suggests the community is interested in the problem. Almost all participating systems beat our (simple) baseline model although there were clear differences in performance between systems.

A variety of approaches to learning from disagreement were also considered, covering nearly all those discussed in the recent survey by Uma et al. (2021b), as well as new ones. Almost all systems trained directly from data containing disagreement, most notably using soft loss training (Uma et al., 2020), including the top 2 systems (`duxy` and `chengcheng`), but also Repeated Labeling (Sheng et al., 2008), instance weighting (Cui, 2023), and the annotator-specific multi-task learning approach recently proposed by Davani et al. (2022). The majority of systems used different methods for training with disagreement for different datasets. A number of systems (e.g., `University at Buffalo`, `SafeWebUH`) used the 'radical perspectivist' approach of training separate models for each annotator (e.g. Akhtar et al., 2021), at least for the tasks with a fixed number of annotators (HS-BREXIT and ArMIS). A few systems used aggregated labels, but not many, with many of the papers arguing that this approach is not appropriate for the tasks at hand. One team (`MIND`) first predicted hard labels trough neural networks and than re-elaborated the prediction space to map the predicted results to a disagreement value. However, it was not just the method that mattered, but the implementation as well: e.g., both the top performing systems and some of the worst performing systems appear to use soft loss training, and there was a substantial difference in performance between the different systems using the approach of Davani et al. (2022), i.e. `iLab` and `morlikowski`.

In terms of architecture, we should mention that although many teams developed different systems for the different datasets, many others took the difference between the datasets as an extra challenge. In particular, 5 systems used multi-task learning for sharing parameters across tasks. One team, `MaChAmp`, took the challenge to the extreme by training a single model for 11 distinct SEMEVAL shared tasks. Mostly variants of BERT were employed, perhaps because our tasks are discriminative rather than generative. Another element emerging from the analysis of the results from the participant systems is the role of additional information in the benchmark data. The systems that leveraged either explicit information about the annotators or extra layers of annotation performed better in general.

### 4.2 Individual datasets results

As already observed for the overall results, also for singular datasets ranking, none of the systems that performed best submitted a paper, but we can still make a few observations. In terms of specific ranking, we can observe a certain degree of variability across datasets, although generally each team ranked similarly across the datasets (average standard deviation is of two positions).

**HS-BREXIT** HS-BREXIT is one of the datasets for which explicit metadata on the annotators is provided, i.e., the demographic group as either muslim immigrant or control group. HS-BREXIT also has a 'dense' annotation, i.e., all annotators annotated every instance. Some systems, e.g., `University at Buffalo`, reported only being able to work in this setting.

`SafeWebUH` (Shahriar and Solorio, 2023), the system ranked 9[th] for this dataset, implemented a system that leverages the annotator metadata, showing an improvement when this information is taken into account. Moreover, `SafeWebUH` found post-aggregation to outperform their alternative soft-label system in terms of cross-entropy score.

The annotator-level information is also lever-aged by the system ranking 10[th] on this dataset (`University at Buffalo`) (Sullivan et al., 2023), in the form of annotator embeddings.

While the main task for this dataset is binary classification on the 'hate' label, the provided extra labels 'offensiveness' and 'aggressiveness' were used by some systems (e.g., IREL Maity et al. 2023), as extra information at the same level as annotator metadata.

| | **HS-BREXIT** | | |
|---|---|---|---|
| | **TEAM** | **CE** | **(REL. $F_1$)** |
| **1** | PALI | 0.235 | *(92.857)* |
| **2** | Duxy | 0.241 | *(92.177)* |
| **2** | king001 | 0.241 | *(92.177)* |
| **4** | chencheng498 | 0.242 | *(93.294)* |
| **5** | ymf924 | 0.251 | *(92.213)* |
| **6** | stce | 0.257 | *(88.798)* |
| **7** | Colain | 0.274 | *(90.888)* |
| **8** | corner | 0.275 | *(92.456)* |
| **9** | SafeWebUH | 0.279 | *(89.713)* |
| **10** | University at Buffalo | 0.295 | *(92.456)* |
| **11** | ccasula | 0.314 | *(88.369)* |
| **12** | sananc | 0.315 | *(89.265)* |
| **13** | Lon-eå | 0.319 | *(88.278)* |
| **14** | eevvgg | 0.328 | *(86.136)* |
| **15** | CICL_DMS | 0.333 | *(86.813)* |
| **16** | Nitrogen_pump | 0.391 | *(84.232)* |
| **17** | UMUTeam | 0.474 | *(90.219)* |
| **18** | IREL | 0.747 | *(86.578)* |
| **19** | iLab | 0.756 | *(55.827)* |
| **20** | MaChAmp | 1.018 | *(86.472)* |
| **21** | MIND | 1.040 | *(78.048)* |
| **22** | babysong | 1.074 | *(85.625)* |
| **23** | xiacui | 2.071 | *(85.599)* |
| **24** | Arguably | 2.686 | *(81.047)* |
| **25** | **Majority Class basel.** | 2.715 | *(84.232)* |
| **26** | morlikowski | 2.992 | *(81.83)* |

Table 4: Results on HS-BREXIT dataset.

**ArMIS** ArMIS was a challenge as the only non-English dataset, meaning that many groups had to use a different LLM for this task (e.g., AraBERT). The most distinctive characteristic of ArMIS from the annotation perspective is that all the data were annotated by three annotators chosen to represent the whole range from conservative to liberal on the political spectrum. For this reason, it made sense to adopt a 'perspectivist' approach to this dataset, modelling the individual annotators and then extracting a soft label from these annotations - as done, e.g., by the Buffalo group (`University at Buffalo`), or Heriot-Watt (`iLab`) (Vitsakis et al., 2023).

The characteristics of ArMIS also provided motivation for the development of the sinusoidal functions approach by team `Lon-eå` (QMUL) (Hosseini et al., 2023).

| **ArMIS** | | |
| Team | CE | (*rel. F1*) |
|---|---|---|
| **1** PALI | 0.469 | *(83.193)* |
| **1** Colain | 0.469 | *(82.084)* |
| **3** chencheng498 | 0.47 | *(83.859)* |
| **3** Duxy | 0.47 | *(83.859)* |
| **5** king001 | 0.472 | *(83.785)* |
| **6** stce | 0.475 | *(83.859)* |
| **7** ymf924 | 0.479 | *(84.653)* |
| **8** sananc | 0.507 | *(81.926)* |
| **9** University at Buffalo | 0.548 | *(77.266)* |
| **10** eevvgg | 0.562 | *(74.279)* |
| **11** SafeWebUH | 0.575 | *(74.918)* |
| **12** ccasula | 0.603 | *(72.097)* |
| **13** CICL_DMS | 0.613 | *(69.506)* |
| **14** Sana | 0.666 | *(84.753)* |
| **14** Lon-eå | 0.666 | *(84.753)* |
| **16** xiacui | 0.685 | *(60.921)* |
| **17** Nitrogen_pump | 0.688 | *(41.676)* |
| **18** MaChAmp | 0.689 | *(47.738)* |
| **19** UMUTeam | 0.713 | *(64.603)* |
| **20** iLab | 1.889 | *(55.373)* |
| **21** IREL | 4.008 | *(56.005)* |
| **22** MIND | 7.845 | *(49.015)* |
| **23** **Majority Class basel.** | 8.908 | *(41.676)* |
| **24** babysong | 11.595 | *(59.416)* |

Table 5: Results on the ArMIS dataset.

| **ConvAbuse** | | |
| Team | CE | (*rel. F1*) |
|---|---|---|
| **1** Duxy | 0.185 | *(94.157)* |
| **2** chencheng498 | 0.186 | *(94.401)* |
| **3** king001 | 0.187 | *(94.401)* |
| **4** stce | 0.189 | *(94.271)* |
| **5** Colain | 0.193 | *(94.271)* |
| **5** PALI | 0.193 | *(94.433)* |
| **7** ymf924 | 0.214 | *(93.871)* |
| **8** sananc | 0.224 | *(91.673)* |
| **9** CICL_DMS | 0.225 | *(92.586)* |
| **10** University at Buffalo | 0.229 | *(91.111)* |
| **11** Lon-eå | 0.234 | *(93.021)* |
| **12** SafeWebUH | 0.236 | *(93.223)* |
| **13** nasheedyasin | 0.242 | *(91.758)* |
| **14** eevvgg | 0.253 | *(91.969)* |
| **15** ccasula | 0.257 | *(93.113)* |
| **16** UMUTeam | 0.302 | *(91.971)* |
| **17** xiacui | 0.318 | *(81.437)* |
| **18** babysong | 0.329 | *(88.611)* |
| **19** Nitrogen_pump | 0.419 | *(74.09)* |
| **20** MaChAmp | 0.474 | *(93.163)* |
| **21** iLab | 0.497 | *(76.844)* |
| **22** dragonfly_captain | 0.788 | *(91.679)* |
| **23** MIND | 1.137 | *(86.513)* |
| **24** **Majority Class basel.** | 3.484 | *(74.09)* |

Table 6: Results on the ConvAbuse dataset.

**ConvAbuse** The best two systems by both metrics (Duxy, chencheng498) used multi-task learning and fine-tuning pretrained models (BERT, DeBERTa). But another team (iLab) used similar methods (fine-tuning MLT) with poorer performance, which they put down to their approach to loss calculation, aimed at preserving all annotators' perspectives at inference time.

ConvAbuse is the only dataset with dialogue context (agent and user turns). Some report using full dialogues (eevvgg by Gajewska (2023), University at Buffalo by Sullivan et al. (2023)) and others only the target final user turn (iLab by Vitsakis et al. (2023)). Most, including all the best performing systems, do not provide this information.

Some teams discuss not being able to use their primary proposed innovations due to the fact that not all annotators label all items (e.g., SafeWebUH). University at Buffalo, was one such team, and developed a distinct method for this dataset. This consisted of adding annotator embeddings and modelling (supposed) interactions and discussions between annotators with cross multi-head attention layers (Shahriar and Solorio, 2023). They showed that this produced a significant performance boost on both metrics compared to a model without this modeling.

**MD-Agreement** At the top of the leader board, 3 systems are tied (Duxy, chencheng498 and king001). Unluckily information available for the systems is scarce as they did not submit a full report, although two of them used multi-task learning and fine-tuning pretrained models.

In terms of leveraging individual annotators behavior, the MD-Agreement dataset has been collected via AMT, and consists of >50,000 annotations form a high number of different annotators (> 800) that are sparsely represented and for which the only information available is the (anonymous) ID of the rater. This poses challenges and requires implementation of a different strategy with respect to the other datasets, leading several teams to disregard this information for this dataset. Some interesting approaches have been used though. The eevvgg (Gajewska, 2023) team characterized each annotator with respect to the answers the others gave, modelling reliability of individual raters against the opinion of the majority. Moreover, University at Buffalo (Sullivan et al., 2023) used annotators agreement to calculate what they call 'polarization' of a topic (there are three topics present in the dataset), i.e. how much a certain topic is judged uniformly or dividing the annotators. Interestingly, these two teams that leveraged the annotator information, positioned well in the MD-Agreement leader board (6[th] and 7[th]), and for both teams this comprised their best ranking obtained among the four datasets.

| | **MD-AGREEMENT** | | |
|---|---|---|---|
| | **Team** | **CE** | **(rel. $F_1$)** |
| 1 | chencheng498 | 0.472 | (84.606) |
| 1 | Duxy | 0.472 | (84.541) |
| 1 | king001 | 0.472 | (84.564) |
| 4 | Colain | 0.475 | (84.616) |
| 5 | PALI | 0.478 | (84.714) |
| 6 | eevvgg | 0.5 | (81.291) |
| 7 | nasheedyasin | 0.508 | (80.987) |
| 7 | SafeWebUH | 0.508 | (82.328) |
| 7 | University at Buffalo | 0.508 | (80.987) |
| 10 | ymf924 | 0.514 | (84.376) |
| 11 | stce | 0.516 | (84.046) |
| 12 | Sana | 0.517 | (82.912) |
| 13 | Lon-eå | 0.521 | (80.329) |
| 14 | sananc | 0.526 | (80.645) |
| 15 | CICL_DMS | 0.531 | (78.933) |
| 16 | MaChAmp | 0.607 | (81.683) |
| 16 | iLab | 0.607 | (74.278) |
| 18 | rana1998al_essa | 0.61 | (62.527) |
| 18 | omaimah | 0.61 | (62.527) |
| 20 | Nitrogen_pump | 0.648 | (73.799) |
| 21 | ccasula | 0.663 | (81.417) |
| 22 | xiacui | 0.694 | (58.007) |
| 23 | UMUTeam | 0.729 | (80.908) |
| 24 | MIND | 2.44 | (71.177) |
| 25 | babysong | 2.665 | (79.316) |
| 26 | **Majority Class basel.** | **7.385** | **(53.375)** |

Table 7: Results on the MD-Agreement dataset.

### 4.3 Selected systems overview

In this Section we briefly highlight some interesting ideas introduced by the participant systems.

**University of Buffalo** (Sullivan et al., 2023) tried three different approaches for the four tasks, based on an analysis of the characteristics of the datasets. We already discussed their approaches for the two datasets in which not all annotators labelled all items, MD-Agreement and ConvAbuse. For HS-Brexit and ArMIS, separate models were trained for each annotator, then combined using ensemble methods.

**Lon-eå** (Hosseini et al., 2023) tested three approaches to predict the soft label: besides a normal sigmoid function, they also tried a *sinusoidal function* and a *step function* taking the number of annotators into account. The intuition was that the possible soft values depend on number of annotators: e.g., in ArMIS, with 3 annotators, the probability of a label can only take the values 0.33, 0.66, and 1. Unfortunately the sinusoidal function approach turned out to not work very well, except for hard evaluation on ArMIS, where the system obtained first place.

**iLab** Vitsakis et al. (2023) did not place highly on the leaderboards, but were among the teams that took the most strongly 'perspectivist' approaches (Cabitza et al., 2023). Their method focused on preserving the distinction between each annotator throughout optimisation and prediction by fine-tuning a multi-task model in which they predicted each annotator's labels with a separate classification layer. They argue that this type of approach avoids the potential for minoritised annotator voices to be ignored or subsumed when optimising to aggregated or distributional metrics.

### 4.4 Comparison and limitations of $F_1$ and CE

Although we believe that a soft form of evaluation is clearly more appropriate for subjective tasks such as in our campaign, we do provide the leader board according to the more canonical hard evaluation $F_1$ metric (Table 8 in the Appendix), for a comparison. Comparing the rankings in Table 8 with those for CE (Table 3) we can see that they are not very dissimilar, but there are some differences. To better understand the origin of these differences, for each of the four datasets, we split the test predictions submitted the 14 best submitted systems into three subsets of data, divided according to three levels of annotators' agreement (low, mid, high)[5] and calculated separate $F_1$. The results in Figure 1 shows a dramatic drop in $F_1$ performance with the decrease of agreement. It is interesting to see how with highly uncertain, or 'difficult' items - the items with an agreement between 40% and 60% - the hard label produced by a system is wrong 46% of the times. Moreover, we observed a moderate correlation between $F_1$ and CE scores calculated on high- and mid-agreement subsets, whereas only a weak correlation exists between CE and $F_1$ scores on data with low agreement.

We also can observe how learning from disagreement in training improved performance in terms of CE with respect to $F_1$. For example, the performance of UMUTEAM that did not include disagreement in training and ranked 13 in $F_1$, while dropping 6 positions in CE ranking. Or the performance of eevvgg and University at Buffalo which obtained their best performance in MD-Agreement

---

[5]High-agreement: more than 83.3% agreement on the same label, either "0" or "1". Mid-agreement: 66.6% to 83.3% agreement, either on "0" or "1" label. Low-agreement: 33.3% to 66.6% agreement on both labels

dataset, improving their average position with respect to their rankings for the other datasets, likely due to the fact that for this dataset they were the only team using information about annotators. All in all, these observations suggest that the observed differences between $F_1$ and CE team rankings are likely to arise from the best ability of CE to represent data with disagreements and how CE is a more appropriate measure than $F_1$ to include disagreements observations in subjective NL tasks, despite some intrinsic limitations.

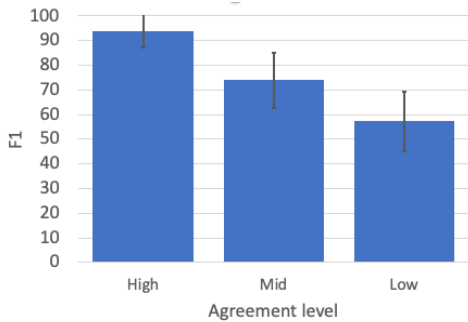

Figure 1: Average F1 calculated separately for different levels of agreement

One observation emerging from Table 3 is that systems generally obtained a higher CE score with ArMIS and MD-Agreement, and a lower CE score with HS-Brexit and ConvAbuse. As pointed out also by team MIND (Rizzi et al., 2023) in an in-depth discussion about evaluation metrics, this depends on the intrinsic entropy of the agreement distribution of each dataset, which in our case is higher for ArMIS and MD-Agreement. Indeed CE can be as low as the total entropy of the distribution. As a side consequence of this behavior, CE scores are not directly comparable across different datasets. Moreover, the relationship between the real distribution and the predicted one, as measured with CE, depends from the distance between them but, since CE is asymmetrical, it depends as well on the specific values involved, thus reflecting only partially the quality of the predictions.

## 5 Conclusions

In this shared task we proposed a framework for learning from disagreements between multiple annotators across a variety of datasets and subjective tasks. We evaluated the ability to predict the distribution of different interpretations of the annotators so to include different point of views.

We are delighted that this second LeWiDi shared task attracted the attention of a substantial number of research groups, and that the participants engaged so actively with the issues raised by the datasets, coming up with such creative solutions. We believe that proposing the same task on a multiplicity of datasets challenged participants and promoted constructive and fruitful thinking as it forced visualisation of the problem in its entirety rather then focusing on the solution to a specific dataset.

One hope is that the shared task and the datasets we released will stimulate further research in this area, by the participant groups and others. To promote this, the Codalab page will remain open to submissions after the deadline so that researchers can continue test their models on the datasets. We believe for instance that further thinking is still needed on issues such as the most appropriate form of evaluation for tasks in which human subjects frequently disagree, or the usefulness of modelling individual annotators or groups of annotators.

## Limitations

The main limitation of this paper is that we couldn't provide a full analysis of the results due to the partial or lacking descriptions for some of the submitted systems, and in particular the best performing ones. Nonetheless, we believe some interesting conclusions could still be reached both about the datasets used and the results obtained. Another limitation of this campaign is that only textual datasets were used, which didn't give us the opportunity to consider the applicability of methods across modalities–a shame, given that this is an area where fruitful interchange has taken place between research in NLP and in computer vision. We hope this limitation will be overcome in future editions, while still encouraging widespread participation.

## Acknowledgments

Elisa Leonardelli was partially supported by the StandByMe European project (REC-RDAP-GBV-AG-2020) on "Stop online violence against women and girls by changing attitudes and behaviour of young people through human rights education" (GA 101005641) and by the StandByMe 2.0 project (CERV-2021-DAPHNE) on "Stop gender-based violence by addressing masculinities and changing behaviour of young people through human rights education" (GA 101049386). Alexandra

Uma and Massimo Poesio were partially supported by the DALI project, ERC Advanced Grant 695662. Massimo Poesio was also partially supported by the ARCIDUCA project, EPSRC grant number EP/W001632/1. Gavin Abercrombie and Verena Rieser were supported by the EPSRC projects 'Gender Bias in Conversational AI' (EP/T023767/1) and 'Equally Safe Online' (EP/W025493/1), and Verena Rieser was also supported by 'AISEC: AI Secure and Explainable by Construction' (EP/T026952/1). Valerio Basile was partially supported by the project "Toxic Language Understanding in Online Communication - BREAKHhateDOWN" funded by Compagnia di San Paolo (ex-post 2020). Barbara Plank is partially supported by the DIALECT project, ERC Consolidator Grant 101043235.

# References

Gavin Abercrombie, Valerio Basile, Sara Tonelli, Verena Rieser, and Alexandra Uma, editors. 2022. *Proceedings of the 1st Workshop on Perspectivist Approaches to NLP @LREC2022*. European Language Resources Association, Marseille, France.

Sohail Akhtar, Valerio Basile, and Viviana Patti. 2021. Whose opinions matter? Perspective-aware models to identify opinions of hate speech victims in abusive language detection. *CoRR*, abs/2106.15896.

Dina Almanea and Massimo Poesio. 2022. ArMIS - the Arabic misogyny and sexism corpus with annotator subjective disagreements. In *Proceedings of the Thirteenth Language Resources and Evaluation Conference*, pages 2282–2291, Marseille, France. European Language Resources Association.

Maria Anzovino, Elisabetta Fersini, and Paolo Rosso. 2018. Automatic identification and classification of misogynistic language on twitter. In *Natural Language Processing and Information Systems*, pages 57–64, Cham. Springer International Publishing.

Lora Aroyo and Chris Welty. 2015. Truth is a lie: Crowd truth and the seven myths of human annotation. *AI Magazine*, 36(1):15–24.

Valerio Basile. 2020. It's the end of the gold standard as we know it. On the impact of pre-aggregation on the evaluation of highly subjective tasks. In *Proc. of the AIXIA Workshop*. Universitá di Torino.

Valerio Basile, Michael Fell, Tommaso Fornaciari, Dirk Hovy, Silviu Paun, Barbara Plank, Massimo Poesio, and Alexandra Uma. 2021. We need to consider disagreement in evaluation. In *Proceedings of the 1st Workshop on Benchmarking: Past, Present and Future*, pages 15–21, Online. Association for Computational Linguistics.

Beata Beigman Klebanov and Eyal Beigman. 2009. From annotator agreement to noise models. *Computational Linguistics*, 35(4):495–503.

Federico Cabitza, Andrea Campagner, and Valerio Basile. 2023. Toward a perspectivist turn in ground truthing for predictive computing. In *Proceedings of the AAAI Conference on Artificial Intelligence*.

Amanda Cercas Curry, Gavin Abercrombie, and Verena Rieser. 2021. ConvAbuse: Data, analysis, and benchmarks for nuanced abuse detection in conversational AI. In *Proceedings of the 2021 Conference on Empirical Methods in Natural Language Processing*, pages 7388–7403, Online and Punta Cana, Dominican Republic. Association for Computational Linguistics.

Xia Cui. 2023. xiacui at SemEval-2023 Task 11: Learning a Model in Mixed-Annotator Datasets using Annotator Ranking Scores as Training Weights. In *Proceedings of the 17th International Workshop on Semantic Evaluation*, Toronto, Canada. Association for Computational Linguistics.

Aida Mostafazadeh Davani, Mark Díaz, and Vinodkumar Prabhakaran. 2022. Dealing with disagreements: Looking beyond the majority vote in subjective annotations. *Transactions of the ACL*, 10:92–110.

Anca Dumitrache, Lora Aroyo, and Chris Welty. 2019. A crowdsourced frame disambiguation corpus with ambiguity. In *Proceedings of the 2019 Conference of the North American Chapter of the Association for Computational Linguistics*, volume 1, pages 2164–2170. Association for Computational Linguistics.

Tommaso Fornaciari, Alexandra Uma, Silviu Paun, Barbara Plank, Dirk Hovy, and Massimo Poesio. 2021. Beyond black & white: Leveraging annotator disagreement via soft-label multi-task learning. In *Proceedings of the 2021 Conference of the North American Chapter of the Association for Computational Linguistics: Human Language Technologies*, pages 2591–2597, Online. Association for Computational Linguistics.

Tommaso Fornaciari, Alexandra Uma, Massimo Poesio, and Dirk Hovy. 2022. Hard and soft evaluation of NLP models with BOOtSTrap SAmpling - BooStSa. In *Proceedings of the 60th Annual Meeting of the Association for Computational Linguistics: System Demonstrations*, pages 127–134, Dublin, Ireland. Association for Computational Linguistics.

Ewelina Gajewska. 2023. eevvgg at SemEval-2023 Task 11: Offensive Language Classification with Rater-based Information. In *Proceedings of the 17th International Workshop on Semantic Evaluation*, Toronto, Canada. Association for Computational Linguistics.

José Antonio García-Díaz, Ronghao Pan, Gema Alcaráz-Marbol, and Maríaand Valencia-García Rafael Marín-Pérez. 2023. UMUTeam at SemEval-2023 Task 11: Ensemble Learning applied to Binary Supervised Classifiers with disagreements. In *Proceedings of the*

*17th International Workshop on Semantic Evaluation (SemEval-2023)*, Toronto, Canada. Association for Computational Linguistics.

Rob van der Goot. 2023. MaChAmp at SemEval-2023 Task 2, 3, 4, 5, 6, 7, 8, 9, 10, 11, and 12: On the effectiveness of intermediate training on an uncurated collection of datasets. In *Proceedings of the 17th International Workshop on Semantic Evaluation (SemEval-2023)*. Association for Computational Linguistics.

Mitchell L. Gordon, Kaitlyn Zhou, Kayur Patel, Tatsunori Hashimoto, and Michael S. Bernstein. 2021a. The disagreement deconvolution: Bringing machine learning performance metrics in line with reality. In *Proceedings of the CHI Conference on Human Factors in Computing Systems*, pages 1–14. ACM.

Mitchell L. Gordon, Kaitlyn Zhou, Kayur Patel, Tatsunori Hashimoto, and Michael S. Bernstein. 2021b. The disagreement deconvolution: Bringing machine learning performance metrics in line with reality. In *Proceedings of the 2021 CHI Conference on Human Factors in Computing Systems*, CHI '21, New York, NY, USA. Association for Computing Machinery.

Dennis Grötzinger, Simon Heuschkel, and Matthias Drews. 2023. CICL_DMS at SemEval-2023 Task 11: Learning With Disagreements (Le-Wi-Di). In *Proceedings of the 17th International Workshop on Semantic Evaluation*, Toronto, Canada. Association for Computational Linguistics.

Peyman Hosseini, Mehran Hosseini, Sana Sabah Al-Azzawi, Marcus Liwicki, Ignacio Castro, and Matthew Purver. 2023. Lon-eå at SemEval-2023 Task 11: A Comparison of Activation Functions for Soft and Hard Label Prediction. In *Proceedings of the 17th International Workshop on Semantic Evaluation*, Toronto, Canada. Association for Computational Linguistics.

Kian Kenyon-Dean, Eisha Ahmed, Scott Fujimoto, Jeremy Georges-Filteau, Christopher Glasz, Barleen Kaur, Auguste Lalande, Shruti Bhanderi, Robert Belfer, Nirmal Kanagasabai, Roman Sarrazingendron, Rohit Verma, and Derek Ruths. 2018. Sentiment analysis: It's complicated! In *Proceedings of the 2018 Conference of the North American Chapter of the Association for Computational Linguistics*, volume 1, pages 1886–1895. Association for Computational Linguistics.

Guneet Singh Kohli and Vinayak Tiwari. 2023. Arguably at SemEval-2023 Task 11: Learning the disagreements using unsupervised behavioral clustering and language models. In *Proceedings of the 17th International Workshop on Semantic Evaluation*, Toronto, Canada. Association for Computational Linguistics.

Elisa Leonardelli, Stefano Menini, Alessio Palmero Aprosio, Marco Guerini, and Sara Tonelli. 2021. Agreeing to disagree: Annotating offensive language datasets with annotators' disagreement. In *Proceedings of the 2021 Conference on Empirical Methods in Natural Language Processing*, pages 10528–10539, Online and Punta Cana, Dominican Republic. Association for Computational Linguistics.

Ankita Maity, Pavan Kandru, Bhavyajeet Singh, Aditya Hari, and Vasudeva Varma. 2023. IREL at SemEval-2023 Task 11: User Conditioned Modelling for Toxicity Detection in Subjective Tasks. In *Proceedings of the 17th International Workshop on Semantic Evaluation*, Toronto, Canada. Association for Computational Linguistics.

Héctor Martínez Alonso, Anders Johannsen, and Barbara Plank. 2016. Supersense tagging with inter-annotator disagreement. In *Proceedings of the 10th Linguistic Annotation Workshop*, pages 43–48. Association for Computational Linguistics.

Rebecca J. Passonneau, Vikas Bhardwaj, Ansaf Salleb-Aouissi, and Nancy Ide. 2012. Multiplicity and word sense: evaluating and learning from multiply labeled word sense annotations. *Language Resources and Evaluation*, 46(2):219–252.

Ellie Pavlick and Tom Kwiatkowski. 2019. Inherent disagreements in human textual inferences. *Transactions of the Association for Computational Linguistics*, 7:677–694.

Joshua C. Peterson, Ruairidh M. Battleday, Thomas L. Griffiths, and Olga Russakovsky. 2019. Human uncertainty makes classification more robust. In *Proceedings of the 2019 IEEE/CVF International Conference on Computer Vision*, pages 9616–9625.

Barbara Plank. 2022. The "problem" of human label variation: On ground truth in data, modeling and evaluation. In *Proceedings of the 2022 Conference on Empirical Methods in Natural Language Processing*, pages 10671–10682, Abu Dhabi, United Arab Emirates. Association for Computational Linguistics.

Barbara Plank, Dirk Hovy, and Anders Søgaard. 2014a. Learning part-of-speech taggers with inter-annotator agreement loss. In *Proceedings of the 14th Conference of the European Chapter of the Association for Computational Linguistics*, pages 742–751. Association for Computational Linguistics.

Barbara Plank, Dirk Hovy, and Anders Søgaard. 2014b. Linguistically debatable or just plain wrong? In *Proceedings of the 52nd Annual Meeting of the Association for Computational Linguistics (Volume 2: Short Papers)*, pages 507–511, Baltimore, Maryland. Association for Computational Linguistics.

Massimo Poesio and Ron Artstein. 2005. The reliability of anaphoric annotation, reconsidered: Taking ambiguity into account. In *Proceedings of the Workshop on Frontiers in Corpus Annotations II: Pie in the Sky*, pages 76–83. Association for Computational Linguistics.

Massimo Poesio, Jon Chamberlain, Silviu Paun, Juntao Yu, Alexandra Uma, and Udo Kruschwitz. 2019. A crowdsourced corpus of multiple judgments and disagreement on anaphoric interpretation. In *Proceedings of the 2019 Conference of the North American Chapter of the Association for Computational Linguistics*, pages 1778–1789. Association for Computational Linguistics.

Marta Recasens, Ed Hovy, and M. Antònia Martí. 2011. Identity, non-identity, and near-identity: Addressing the complexity of coreference. *Lingua*, 121(6):1138–1152.

Giulia Rizzi, Alessandro Astorino, Daniel Scalena, Paolo Rosso, and Elisabetta Fersini. 2023. MIND at SemEval-2023 Task 11: From Uncertain Predictions to Subjective Disagreement. In *Proceedings of the 17th International Workshop on Semantic Evaluation*, Toronto, Canada. Association for Computational Linguistics.

Filipe Rodrigues and Francisco C. Pereira. 2018. Deep learning from crowds. In *Proceedings of the 32nd AAAI Conference on Artificial Intelligence*, pages 1611–1618.

Marta Sandri, Elisa Leonardelli, Sara Tonelli, and Elisabetta Jezek. 2023. Why Don't You Do It Right? Analysing Annotators' Disagreement in Subjective Tasks. In *Proceedings of the 2023 Conference of the European Chapter of the Association for Computational Linguistics*.

Sadat Shahriar and Thamar Solorio. 2023. SafeWebUH at SemEval-2023 Task 11: Learning Annotator Disagreement in Derogatory Text: Comparison of Direct Training vs Aggregation. In *Proceedings of the 17th International Workshop on Semantic Evaluation*, Toronto, Canada. Association for Computational Linguistics.

Victor S. Sheng, Foster Provost, and Panagiotis G. Ipeirotis. 2008. Get another label? Improving data quality and data mining using multiple, noisy labelers. In *Proceedings of the 14th ACM SIGKDD International Conference on Knowledge Discovery and Data Mining*, pages 614–622.

Edwin Simpson, Erik-Lân Do Dinh, Tristan Miller, and Iryna Gurevych. 2019. Predicting humorousness and metaphor novelty with Gaussian process preference learning. In *Proceedings of the 57th Annual Meeting of the Association for Computational Linguistics*, pages 5716–5728. Association for Computational Linguistics.

Michael J. Sullivan, Mohammed Nasheed Yasin, and Cassandra L. Jacobs. 2023. University at Buffalo at SemEval-2023 Task 11: MASDA–Modelling Annotator Sensibilities through DisAggregation. In *Proceedings of the 17th International Workshop on Semantic Evaluation*, Toronto, Canada. Association for Computational Linguistics.

Alexandra Uma, Dina Almanea, and Massimo Poesio. 2022. Scaling and disagreements: Bias, noise and ambiguity. *Frontiers in Artificial Intelligence: Human-Centered AI*.

Alexandra Uma, Tommaso Fornaciari, Anca Dumitrache, Tristan Miller, Jon Chamberlain, Barbara Plank, Edwin Simpson, and Massimo Poesio. 2021a. SemEval-2021 task 12: Learning with disagreements. In *Proceedings of the 15th International Workshop on Semantic Evaluation (SemEval-2021)*, pages 338–347, Online. Association for Computational Linguistics.

Alexandra Uma, Tommaso Fornaciari, Dirk Hovy, Silviu Paun, Barbara Plank, and Massimo Poesio. 2020. A case for soft-loss functions. In *Proceedings of the 8th AAAI Conference on Human Computation and Crowdsourcing*, pages 173–177.

Alexandra Uma, Tommaso Fornaciari, Dirk Hovy, Silviu Paun, Barbara Plank, and Massimo Poesio. 2021b. Learning from disagreement: A survey. *Journal of Artificial Intelligence Research*, 72:1385–1470.

Yannick Versley. 2008. Vagueness and referential ambiguity in a large-scale annotated corpus. *Research on Language and Computation*, 6(3):333–353.

Nikolas Vitsakis, Amit Parekh, Tanvi Dinkar, Gavin Abercrombie, Ioannis Konstas, and Verena Rieser. 2023. iLab at SemEval-2023 Task 11 Le-Wi-Di: Modelling Disagreement or Modelling Perspectives? In *Proceedings of the 17th International Workshop on Semantic Evaluation*, Toronto, Canada. Association for Computational Linguistics.

Ruyuan Wan and Karla Badillo-Urquiola. 2023. Dragonfly_captain at SemEval-2023 Task 11: Unpacking Disagreement with Investigation of Annotator Demographics and Task Difficulty. In *Proceedings of the 17th International Workshop on Semantic Evaluation*, Toronto, Canada. Association for Computational Linguistics.

Aaron Steven White, Rachel Rudinger, Kyle Rawlins, and Benjamin Van Durme. 2018. Lexicosyntactic inference in neural models. In *Proceedings of the 2018 Conference on Empirical Methods in Natural Language Processing*, pages 4717–4724, Brussels, Belgium. Association for Computational Linguistics.

# Appendix

| | | | HARD EVALUATION | | | | | | | |
|---|---|---|---|---|---|---|---|---|---|---|
| | | | HS-Brexit | | ArMIS | | ConvAbuse | | MD-Agr | |
| Rank | (av.pos) | Team | micro-$F_1$ | (pos) | micro-$F_1$ | (pos) | micro-$F_1$ | (pos) | micro-$F_1$ | (pos) |
| 1 | (2.25) | chengcheng498 | 93.294 | (1) | 83.859 | (5) | 94.932 | (1) | 84.618 | (2) |
| 2 | (3.50) | Duxy | 93.166 | (2) | 83.859 | (5) | 94.818 | (2) | 84.541 | (5) |
| 2 | (3.50) | PALI | 92.857 | (3) | 83.859 | (5) | 94.433 | (5) | 84.714 | (1) |
| 4 | (4.75) | ymf924 | 92.458 | (6) | 84.653 | (3) | 94.69 | (3) | 84.376 | (7) |
| 4 | (4.75) | stce | 92.668 | (5) | 83.984 | (4) | 94.48 | (4) | 84.525 | (6) |
| 6 | (6.00) | Colain | 91.667 | (9) | 83.859 | (5) | 94.271 | (7) | 84.616 | (3) |
| 7 | (6.75) | king001 | 92.177 | (8) | 83.785 | (9) | 94.401 | (6) | 84.564 | (4) |
| 8 | (10.50) | SafeWebUH | 89.713 | (12) | 74.918 | (12) | 93.223 | (8) | 82.328 | (10) |
| 9 | (11.00) | ccasula | 90.695 | (10) | 72.097 | (14) | 93.113 | (11) | 82.78 | (9) |
| 9 | (11.00) | Lon-eâ | 88.278 | (14) | 84.753 | (1) | 93.021 | (12) | 80.329 | (17) |
| 11 | (12.00) | Univ. at Buffalo | 92.456 | (7) | 77.266 | (11) | 91.758 | (16) | 80.987 | (14) |
| 12 | (12.25) | sananc | 89.617 | (13) | 81.926 | (10) | 93.143 | (10) | 80.651 | (16) |
| 13 | (13.50) | UMUTeam | 90.219 | (11) | 68.741 | (16) | 92.458 | (14) | 81.467 | (13) |
| 13 | (13.50) | Sana | - | (22) | 84.753 | (1) | - | (23) | 82.912 | (8) |
| 15 | (14.25) | MaChAmp | 87.205 | (15) | 47.738 | (21) | 93.163 | (9) | 81.683 | (12) |
| 15 | (14.25) | eevvgg | 86.136 | (18) | 74.279 | (13) | 91.969 | (15) | 81.866 | (11) |
| 17 | (15.75) | CICL_DMS | 86.813 | (16) | 69.506 | (15) | 92.586 | (13) | 78.933 | (19) |
| 18 | (18.50) | babysong | 85.625 | (19) | 59.416 | (18) | 88.611 | (19) | 79.316 | (18) |
| 19 | (19.00) | nasheedyasin | - | (22) | - | (24) | 91.758 | (16) | 80.987 | (14) |
| 20 | (19.25) | corner | 92.737 | (4) | - | (24) | - | (23) | - | (26) |
| 21 | (20.75) | xiacui | 85.599 | (20) | 60.921 | (17) | 81.437 | (21) | 58.007 | (25) |
| 22 | (21.25) | MIND | 85.045 | (21) | 49.015 | (22) | 86.745 | (20) | 75.396 | (22) |
| 22 | (21.25) | IREL | 86.578 | (17) | 56.005 | (19) | - | (23) | 37.407 | (26) |
| 24 | (21.75) | iLab | 84.227 | (22) | 55.373 | (20) | 76.844 | (22) | 74.278 | (23) |
| 25 | (22.25) | rana1998al_essa | - | (22) | - | (24) | - | (23) | 77.89 | (20) |
| 25 | (22.25) | omaimah | - | (22) | - | (24) | - | (23) | 77.89 | (20) |
| 27 | (22.50) | dragonfly_captain | - | (22) | - | (24) | 91.679 | (18) | - | (26) |
| 28 | (23.00) | Nitrogen_pump | 84.232 | (22) | 47.229 | (23) | 74.09 | (23) | 73.799 | (24) |
| 29 | (23.75) | **Majority Class basel.** | 84.232 | (22) | 41.676 | (24) | 74.09 | (23) | 53.375 | (26) |
| 29 | (23.75) | morlikowski | 81.83 | (22) | - | (24) | - | (23) | - | (26) |
| 29 | (23.75) | Arguably | 81.047 | (22) | - | (24) | - | (23) | - | (26) |

Table 8: Overall hard evaluation results as an average of a system's position across datasets

## A  $F_1$ results

The general leader board according to $F_1$ scores is shown in Table 8. For each team, the best $F_1$ among submitted predictions are shown (each team could submit up to 5 predictions for dataset). Thus they can differ from the $F_1$ shown in Tables 5, 6, 4 and 7, which are the "relative $F_1$", i.e. relative to the systems with the lowest CE.

## B  Error Analysis

We based the error analyses of soft evaluation on the 14 best performing systems because their errors are the most indicative regarding difficulties in solving our task (but also because those teams submitted predictions for all datasets).

All items present two soft labels ("0" and "1") which represent the agreement level for the each possible label, expressed as the proportion of annotations in favour of one or the other label. By design, for each item, soft labels "0" and "1" are in the [0-1] range and sum up to 1. The possible specific ratios between agreement/disagreement that a soft label can assume for each item, depends in first instance on the number of annotations collected for the item: for example, given an odd number of annotations, soft labels cannot be equal ["0": 0.5, "1": 0.5]. Hence, since the 4 datasets employed distinct numbers of annotators, they exhibits not homogeneous discrete distributions of agreement levels. To be able to generalize across levels of agreement between annotators and across datasets, we thus split the range of possible combination of judgments into 5 bins, [6] and each item of all datasets has been accordingly reassigned to one of the 5 bins. For example the bin [1-.8] includes all the items for which 80% to 100% of annotators agreed on label "0". For simplicity, we refer to each bin using only the probability of label "0" since probability of label "1" is the remaining difference.

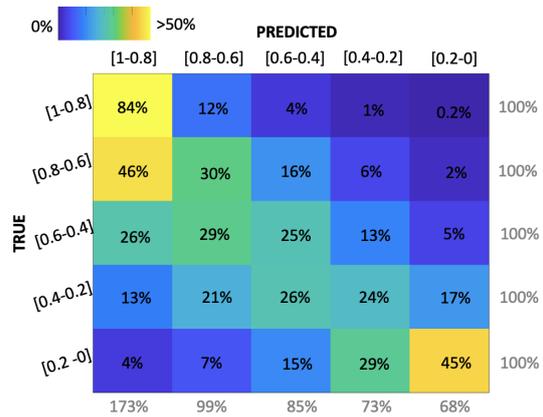

Figure 2: Soft labels confusion matrix. For simplicity of representation, only the probability of label "0" are indicated, given that label "1" and "0" probabilities sum is always 1.

Considering the same bins of agreement, we calculated the confusion matrix between true and predicted soft labels shown in Figure 2. The matrix shows how the items with more disagreement (bins comprehending level of agreement between 0.8 to 0.2) are classified less precisely, i.e. the diagonal has lower values for these probabilities which often are classified into adjacent level of disagreement. On the contrary, items with higher agreement are more correctly classified, with 84% for label "0" and 45% for label "1". Furthermore, soft label tends towards label "0", as the total percentages under the predicted columns show.

Finally, the test of the MD-AGREEMENT dataset has been recently further annotated for the presumable reason of disagreement, following a two-levels taxonomy of disagreement (Sandri et al., 2023). For this dataset, it has been shown how only in less than 1% of cases, disagreement is due to sloppiness of the annotators, while around 15% is due to insufficient information contained in the tweet, 20% to ambiguity in text and over 65% is due to subjectivity in judgments. This confirms that, at least for carefully collected annotations, what we observe is genuine disagreement and not only poor annotations' quality. Moreover, we used this second layer of annotations, to understand whether the best systems partecipating in LeWiDi performed particularly good for specific types of disagreement and if some types of disagreements instead where incorrectly classified by all systems. However, at least for the MD-AGREEMENT, from a qualitative analysis we failed to observe a specific pattern of errors' distribution in relationship to the category of disagreements.

---

[6](["0": 1.0-0.8, "1": 0.0-0.2], ["0": 0.8-0.6, "1": 0.2-0.4], ["0": 0.6-0.4, "1": 0.4-0.6], ["0": 0.4-0.2, "1": 0.6-0.8], ["0": 0.2-0.0, "1": 0.8-1.0]